%% file: main.tex
\documentclass[acmlarge,nonacm]{acmart}
\usepackage{amsmath,amsfonts}
\usepackage{algorithmic}
\usepackage{graphicx}
\usepackage{textcomp}
\usepackage{todonotes}
\usepackage[ruled,vlined,linesnumbered]{algorithm2e}
\usepackage{soul}
\usepackage{svg}
\usepackage{caption}
\usepackage{subcaption}
\usepackage{amsmath}
\usepackage{booktabs}
\usepackage{hyperref}
\usepackage{balance}
\usepackage{tabulary}
\usepackage{multicol}
\usepackage{stfloats}
\usepackage{tabularx}
\usepackage{pbox}
\usepackage{footnote}

\newcommand{\citx}[1]{{\citestyle{acmauthoryear}\citealt{#1}} \cite{#1}}
\newcommand{\citt}[1]{{\emph{\citestyle{acmauthoryear} \protect\NoHyper \hspace{-1ex}(\citealt{#1})} \cite{#1}}}

\setcitestyle{numbers}

\AtBeginDocument{%
  \providecommand\BibTeX{{%
    \normalfont B\kern-0.5em{\scshape i\kern-0.25em b}\kern-0.8em\TeX}}}

\begin{document}

\title{BayesBeat: Reliable Atrial Fibrillation Detection from Noisy Photoplethysmography Data}

\author{Sarkar Snigdha Sarathi Das}
\orcid{0000-0003-1052-4142}
\authornote{Equal contribution}
\authornote{Work done while the author was a student of Bangladesh University of Engineering and Technology}
\email{sfd5525@psu.edu} 
\affiliation{%
  \institution{Bangladesh University of Engineering and Technology}
  \city{Dhaka}
  \country{Bangladesh}
}\affiliation{%
  \institution{Pennsylvania State University}
  \country{USA}
}

\author{Subangkar Karmaker Shanto}
\orcid{0000-0003-0394-2565}
\email{subangkar.karmaker@gmail.com}
\authornotemark[1]
\affiliation{%
  \institution{Bangladesh University of Engineering and Technology}
  \country{Bangladesh}
}\affiliation{%
  \institution{United International University}
  \city{Dhaka}
  \country{Bangladesh}
}

\author{Masum Rahman}
\email{masumrahman99@gmail.com}
\orcid{0000-0002-8214-4985}
\affiliation{%
  \institution{Bangladesh University of Engineering and Technology}
  \city{Dhaka}
  \country{Bangladesh}
}

\author{Md Saiful Islam}
\authornotemark[2]
\email{mislam6@ur.rochester.edu}
\orcid{0000-0003-3725-3493}
\affiliation{%
  \institution{Bangladesh University of Engineering and Technology}
  \city{Dhaka}
 \country{Bangladesh}
}
\affiliation{%
    \institution{University of Rochester}
    \city{Rochester}
    \state{New York}
    \country{USA}
}
\author{Atif Hasan Rahman}
\orcid{0000-0003-1805-3971}
\email{atif@cse.buet.ac.bd}
\affiliation{%
  \institution{Bangladesh University of Engineering and Technology}
  \city{Dhaka}
  \country{Bangladesh}
}
\author{Mohammad M. Masud}
\email{m.masud@uaeu.ac.ae}
\orcid{0000-0002-5274-5982}
\affiliation{%
  \institution{United Arab Emirates University }
  \city{Al Ain}
  \country{UAE}
}

\author{Mohammed Eunus Ali}
\orcid{0000-0002-0384-7616}
\authornote{Corresponding author}
\email{eunus@cse.buet.ac.ad}
\affiliation{%
  \institution{Bangladesh University of Engineering and Technology}
  \city{Dhaka}
  \country{Bangladesh}
}

\renewcommand{\shortauthors}{Das and Shanto, et al.}
\begin{abstract}
\input{files/abstract.tex}
\end{abstract}




\maketitle

\section{Introduction}
\label{sec:introduction}
\input{files/introduction}

\section{Related Work}
\label{sec:related}
\input{files/related}

\section{Problem Formulation}
\label{sec:problem}
\input{files/problem}

\section{Our Approach: BayesBeat}
\label{sec:method}
\input{files/methodology}

\section{Experiments}
\label{sec:experiment}
\input{files/results}


\section{Conclusion}
\label{sec:conclusion}
\input{files/conclusion}


\section{Code Availability}
\label{sec:codebase}
\input{files/code_availability}

\section*{Acknowledgments}
\input{files/acknowledgement}

\balance
\bibliographystyle{ACM-Reference-Format}
\bibliography{bibilography}

\end{document}

%% file: files/abstract.tex

Smartwatches or fitness trackers have garnered a lot of popularity as potential health tracking devices due to their affordable and longitudinal monitoring capabilities. To further widen their health tracking capabilities, in recent years researchers have started to look into the possibility of Atrial Fibrillation (AF) detection in real-time leveraging photoplethysmography (PPG) data, an inexpensive sensor widely available in almost all smartwatches. A significant challenge in AF detection from PPG signals comes from the inherent noise in the smartwatch PPG signals. In this paper, we propose a novel deep learning based approach, BayesBeat that leverages the power of Bayesian deep learning to accurately infer AF risks from noisy PPG signals, and at the same time provides an uncertainty estimate of the prediction. Extensive experiments on two publicly available dataset reveal that our proposed method BayesBeat outperforms the existing state-of-the-art methods. Moreover, BayesBeat is substantially more efficient having 40-200X fewer parameters than state-of-the-art baseline approaches making it suitable for deployment in resource constrained wearable devices.


%% file: files/introduction.tex
 {A}{trial} fibrillation (AF) is the most prevalent form of arrhythmia, a type of abnormality characterized by irregular beating of the two upper chambers (the atria) of the heart. AF can often go unnoticed for a long time and can lead to severe complications such as stroke and heart failure if not controlled at an early stage. Hence, early diagnosis of AF can significantly reduce the risk of death from the above complications.
 As most of the AF cases are asymptomatic at initial stages, periodic clinical ECG assessments have a low probability of detecting AF~\cite{pereira2020photoplethysmography}. The increasing popularity of smartwatches (or fitness trackers) equipped with photoplethysmography (PPG) sensors opens a new opportunity for developing non-invasive continuous real-time monitoring of AF from PPG signals.

PPG sensors essentially use reflection intensity of emitted light from body parts to monitor volumetric blood flow alterations, which are ultimately converted to heart rate and heart beating patterns. Proper contact with the body is fundamental to capturing noise-free signals by these body sensors.
However, PPG signals are difficult to process and use for AF prediction than ECG considering the influence of various physiological factors \cite{zhu2021learning}. Moreover, several recent studies show that motion artifacts during activities such as walking, running, etc. can significantly deteriorate the accuracy of heart rate calculation~\cite{tuauctan2015characterization, zhang2019motion}. Furthermore, skin tone
~\cite{nitzan2014pulse, ries1989skin} and fastening belt pressure~\cite{shimazaki2018effect} have also been attributed as sources of noise in PPG signals. 
Hence, both noise artifacts and the existence of AF can cause irregular pulse-to-pulse intervals in PPG signals, which pose a major challenge in detecting AF from the captured signals.

A large body of research in this domain relies on hand-crafted feature extraction from PPG signals to detect AF.
Manual feature extraction requires designing consistent features for diverse PPG environments, across different individuals, while maintaining high discriminatory power~\cite{elgendi2016optimal, torres2020multi}. 
These approaches do not yield reliable performance in many situations as high noise negatively affects the feature extraction phases~\cite{zhang2014troika}.
Inspired by the success of deep learning based techniques in ECG classification, a few recent approaches apply deep convolutional neural network (CNN) for prediction of AF risks from smartwatch PPG data~\cite{pereira2020photoplethysmography, tison2018passive, aliamiri2018deep, shen2019ambulatory}.

~\cite{aliamiri2018deep} collected PPG signals in a controlled environment and filtered out the poor quality signals to increase detection accuracy. As this solution was not suitable for deployment in ambulatory settings, the authors later proposed to use ResNeXt~\cite{xie2017aggregated} architecture without any explicit filtering step~\citt{shen2019ambulatory}. However, since significant portions of PPG signals are noisy, the proposed model fails to accurately infer AF labels for those segments.

\begin{figure*}[ht]
\begin{center}
\captionsetup{justification=centering}
\includegraphics[width=\textwidth]{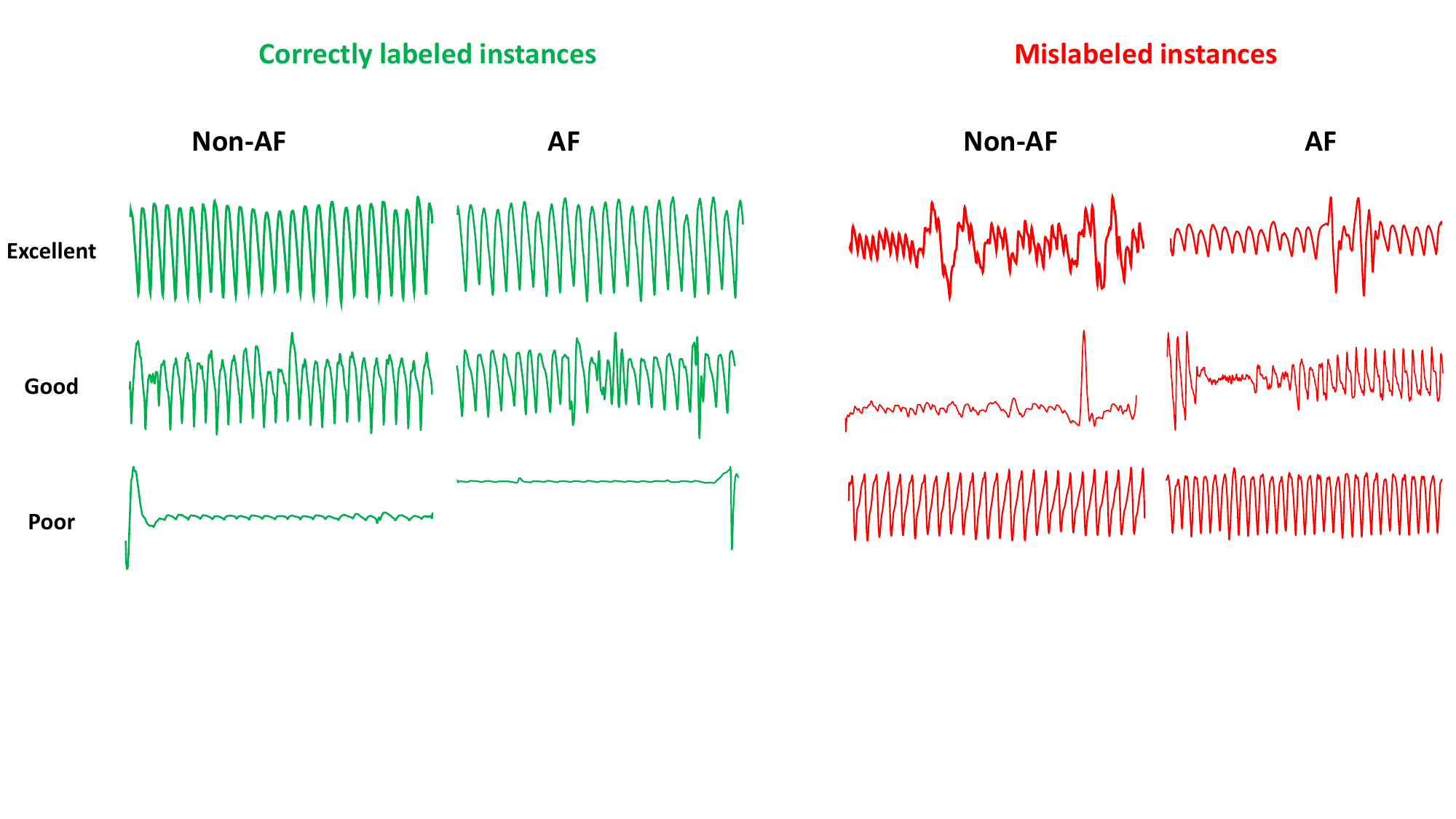}

\vspace{-25mm}
\caption{Samples of PPG signals from the Stanford Wrist Photoplethysmography dataset~\cite{torres2020multi}. Signals on the left side are correctly labeled. Non-AF signals from healthy patients are at first column whereas signals of second columns are from subjects with AF. \cite{torres2020multi}~leveraged a semi-supervised scheme to automatically label the signal quality as Excellent, Good, or Poor. Although many signals are labeled correctly as shown in left, this scheme makes frequent mistakes and labels many signals incorrectly, showed in right side.}

\label{fig:fig_comparesignals}
\end{center}
\end{figure*}

In order to address the issue of noise, ~\cite{elgendi2016optimal} proposed a signal quality index that categorizes PPG signals into three classes: excellent, acceptable (good), and unfit (poor) for diagnosis. In a recent work,~\citt{torres2020multi} proposed a multi-task approach, where  they first estimate the signal quality, and then in the next phase, detect AF with signals that are of excellent qualities. 
This state-of-the-art work has several limitations. They learn signal quality estimation in a supervised manner that requires huge human involvement for quality label annotation. To reduce the human effort, they manually annotate a part of the dataset using \cite{elgendi2016optimal} and then label the rest of the dataset using a separate model trained on the human-labeled dataset. However, this automated labeling approach may generate imprecise quality labels. Figure~\ref{fig:fig_comparesignals} shows some of the signals that are \emph{correctly labeled} and \emph{wrongly labeled}, respectively, as excellent, good, and poor quality.
These inaccurate ground truth quality labels may propagate errors to the downstream model resulting in performance degradation. The proposed model also does not generalize well across the whole dataset, i.e., the results are excellent in test set, however, validation and training set results are not satisfactory. 
These regular deep learning models also cannot give any probabilistic guarantee of prediction as they cannot learn the true variability of the input data.


Summarizing the above prior works, AF detection involves first employing traditional noise filtering techniques~\cite{shen2019ambulatory, lee2003reduction, raghuram2012dual} on the input signal, and subsequently applying different classification methods to categorize them as either AF or non-AF. However, these filtering techniques are often insufficient due to the high degree of noise present in PPG signals. More importantly, signal irregularities in PPG signals can be originated either from external noise or from AF-related arrhythmic irregularities. Thus, applying aggressive filtering schemes may truncate AF-related irregularities as well, leading to suboptimal performance. These shortcomings of the prior works warrant investigating into more sophisticated classification methods robust to noise that can give a probabilistic guarantee of prediction for reliable AF detection.

To overcome the above limitations of handling uncertain PPG sensor data, in this paper, we propose a novel Bayesian deep learning approach, \emph{BayesBeat} to provide a probabilistic guarantee of AF detection. 
Moreover, Bayesian deep learning models learn the distribution of network weights instead of fixed value weights. Thus the rich representations learnt from the input data are robust to perturbations from signal noise. The key intuition of our approach is to exploit the Bayesian deep learning based approach to capture the inherent uncertainty in the noisy PPG signals and take this uncertainty into account while detecting AF from PPG signals, rather than training separate models to assess the signal quality. 
The advantage of  \emph{BayesBeat} is multi-fold:
(i) \textbf{robust:} can handle noisy PPG data and also avoid over-fitting for a small dataset, (ii) \textbf{efficient:} requires much fewer parameters, yet achieves state-of-art performance guarantee, which makes it suitable for being deployed in low-end handheld devices, and (iii) \textbf{uncertainty estimate:} gives aleatoric uncertainty (i.e. stochasticity of data resulting from input noise), which obviates the requirement of any supervision for signal quality estimation as the model itself learns to distinguish between AF related and motion-induced irregularities in PPG signals.

We have conducted extensive experiments with our model and compared it against the state-of-the-art AF detection algorithms with the largest publicly available Stanford wrist photoplethysmography dataset \citt{torres2020multi}. Our proposed model BayesBeat substantially outperforms state-of-the-art AF detection methods achieving 7.1-9.2\% higher F1-score than~\citt{shen2019ambulatory} and 10-20\% higher F1-score than~\citt{torres2020multi}. Additionally, we experiment the generalizability of these trained models in MIMIC-III PPG dataset \cite{goldberger2000physiobank}. We find that BayesBeat is surprisingly effective in generalizing to PPG data from other domains without any additional fine-tuning, outperforming state-of-the-art baselines by 16-20\%  absolute F1-points. Moreover, BayesBeat requires 40-200X fewer parameters than state-of-the-art baseline approaches making it suitable for deployment in low-resource wearable devices. Another important finding is that, \citt{torres2020multi} achieves the best result when it takes the signals with excellent labels only, i.e. rejecting more than 80\% of the data; whereas BayesBeat can handle the noisy data gracefully.

In summary, we have made the following contributions:

\begin{itemize}
    \item We propose a novel deep learning based approach, namely BayesBeat, to automatically detect AF from PPG signals.
    \item We also provide an uncertainty estimation of the prediction, which can be handy in a real-world setting with noisy PPG data.
    \item Our experimental results show that BayesBeat outperforms state-of-the-art methods while requiring a significantly smaller number of parameters compared to them. This makes it highly suitable for deployment in low-end wearable devices.
\end{itemize}

%% file: files/related.tex
Existing works on AF detection from PPG data can be divided into three categories: traditional statistical analysis, classical machine learning (ML) based methods, and deep learning (DL) based methods.

\noindent\textbf{Traditional Statistical Analysis Approaches}: Most of the statistical approaches involved extracting different features from rhythm data, e.g. R to R interval, Shannon entropy, RMSSD, Poincare plot, etc. Each of these features was analyzed extensively to determine thresholds for Atrial Fibrillation detection~\cite{lee2013time, krivoshei2017smart}. To handle noise artifacts in extracted signals, some later works also added movement detection to fix the thresholds \cite{bashar2018developing, chong2016motion}. In addition, various noise reduction techniques for PPG signals with motion artifacts based noises have also been explored~\cite{wang2010artifact, raghuram2012dual, lim2018adaptive, lee2003reduction, lee2007periodic, han2012artifacts}. Nevertheless, these traditional statistical approaches fail to capture complex patterns present in PPG data due to limited representation power. 

\noindent\textbf{Classical Machine Learning based Approaches}: To improve performance over statistical approaches, several machine learning algorithms have been applied to comprehensive features extracted from PPG data. SVM \cite{shan2016reliable, lemay2016wrist}, logistic models \cite{tang2017identification, nemati2016monitoring, schack2017computationally}, KNN~\cite{corino2017detection}, decision trees \cite{fallet2019can}, etc. have replaced traditional threshold-based detection enabling capturing of complex relationship between different parameters. However, the success of these methods relied heavily on elaborate feature engineering by domain experts, which is quite challenging for inherently noisy PPG signals  due to motion artifacts. 

\noindent\textbf{Deep Learning (DL) based Approaches}: Unlike other machine learning methods, deep learning algorithms can learn complex feature representations eliminating the requirement of hand-picked features. Consequently, state-of-the-art AF detection algorithms leverage deep learning for accurate prediction. ~\cite{aliamiri2018deep} proposed a convolutional recurrent hybrid model for AF detection. They also used a separate quality assessment network for estimating signal quality. However, since they only worked with controlled environment data, it was not suitable for usage in an ambulatory noisy setting. To alleviate this problem, in a later work~\citt{shen2019ambulatory}, they used 1D ResNeXt architecture~\cite{xie2017aggregated}, a model with over 7.5 million parameters which they found to be robust against noise artifacts. Nevertheless, a model that considers all signal segments equally is not practical in the real world setting, since a significant portion of signals gets heavily distorted by noise artifacts from micro-movement, uneven skin contact, etc. Furthermore, a computation heavy model like ResNeXt is not suitable to be deployed in consumer devices with limited power (e.g., wearable devices). Later, \cite{poh2018diagnostic} used convolutional neural network for AF prediction. {However, their study was limited to fingertip PPG records in clinical setting, and is not suited for noisy signals from wrist-worn wearable devices}. To assess the problems related to noise, \citt{torres2020multi} recently collected a large dataset involving both AF and non-AF individuals. Besides, for supervised training of signal quality, they labeled each signal segment with corresponding quality index \cite{elgendi2016optimal}. This labeling process is partially done manually which is later utilized by a separate model for labeling the rest of the dataset. However, the automated labeling can be erroneous, and labeling errors can propagate to the final AF detection model leading to suboptimal performance.

%% file: files/problem.tex
Let $X = \{X_k\}, k = 1,2,...,N$ be the set of $N$ PPG signal segments obtained from wearable PPG sensors. Each segment $X_k$ is of length $l$, i.e., $X_k\in \mathbb{R}^l$. Our goal is to predict AF label $Y_k \in \{0,1\}$, which represents negative (Non-AF) or positive (Atrial Fibrillation), for each given $X_k \in X$. We refer to this as the Atrial Fibrillation (AF) detection problem.

A major challenge in AF detection from wearable PPG signals is the inherent noise due to the physical limitation of current wearable technologies and motion artifacts of the carrier. Figure~\ref{fig:fig_comparesignals} shows example signals from the Stanford wearable Photoplethysmography dataset~\citt{torres2020multi}, where we can see Excellent, Good, and Poor signals of both AF and non-AF categories. Existing methods~\citt{torres2020multi} fail to capture these noises and sometimes mis-interpret these noises (Figure~\ref{fig:fig_comparesignals}), which may lead to poor performance in AF detection from noisy signals.

To overcome the above challenges in dealing with the varying degree of noise in input signals, we propose a Bayesian convolutional neural network architecture, \textit{BayesBeat}. The key intuition of \textit{BayesBeat} is to exploit the Bayesian deep learning based approach to capture the inherent uncertainty in the noisy PPG signals 
Predictions on noisy signals are likely to yield high uncertainty scores. Hence the uncertainty score works as a proxy for signal quality, eliminating the need for a separate signal quality estimation model.

%% file: files/methodology.tex
In this section, we first give an overview of Bayesian deep learning. We then present a detailed formulation of Bayesian networks used in BayesBeat followed by the inference procedure. Finally,  we describe our BayesBeat architecture and the training algorithm. 

\subsection{Bayesian Deep Learning} 
Bayesian deep learning has recently shown its promise to capture uncertainty in real-world data by learning posterior distributions of the weights from training data sets. Formally, given a training dataset $\mathcal{D}$, we first learn the posterior distribution of the weights $P(\mathbf{w}|\mathcal{D})$. Then, by using this posterior, the distribution of predicted label $\mathbf{y^{\prime}}$ given unseen test data $\mathbf{x^{\prime}}$, is computed as $P(\mathbf{y^{\prime}} | \mathbf{x^{\prime}}) = \mathbb{E}_{P(\mathbf{w}|\mathcal{D})}[P(\mathbf{y^{\prime}}|\mathbf{x^{\prime}}, \mathbf{w})]$~\cite{blundell2015weight}. However, computing the exact posterior requires the calculation of the probability of evidence $P(\mathcal{D})$, which is computationally intractable due to the high dimensionality of model parameters and real-world data. Hence, several techniques have been proposed for approximating posterior distribution~\cite{blundell2015weight, kingma2015variational, gal2016dropout}. ~\cite{blundell2015weight} proposed a backpropagation compatible algorithm \textit{Bayes by backprop} for approximating the posteriors. Later ~\cite{kingma2015variational} employed a local reparameterization trick to achieve greater efficiency and lower variance gradient estimates. We adapt the above two techniques ~\cite{blundell2015weight, kingma2015variational} to model our posterior distribution and detect AF from PPG signals. Note that, this posterior can also be inferred by using simple Monte Carlo dropout \cite{gal2016dropout}. However, recent studies suggest that this approach lacks expressiveness and does not fully capture the uncertainty in model predictions \cite{jospin2020hands}.

\subsection{Modeling Bayesian Networks for BayesBeat}
Unlike traditional deep learning layers that learn weights as points, all the parameter weights of our network are posterior probability distributions. As a result, the outputs of the network are probabilities from which we can calculate the uncertainty of a decision taken by the network. Figure \ref{fig:fig_archi}(b) shows a high-level representation of how an input is transformed into output distributions by using a 1D Bayesian convolutional layer. As the exact computation of posterior is intractable, our target is to approximate the true posterior $P(\mathbf{w} | \mathcal{D})$. For this, we first learn parameters $\theta$ for computing distribution of weights $q(\mathbf{w} | \theta)$ from which we can approximate $P(\mathbf{w} | \mathcal{D})$. To do this, we minimize the Kullback-Leibler (KL) divergence between $q(\mathbf{w} | \theta)$ and $P(\mathbf{w} | \mathcal{D})$, i.e., the goal is to minimize the cost function $\mathcal{F}(\mathcal{D},\theta) = KL[q(\mathbf{w} | \theta)~||~P(\mathbf{w|\mathcal{D}})] $. After performing algebraic simplifications~\cite{blundell2015weight}, the above minimization target can be expressed as follows:

\begin{equation}
    \label{eq:cost}
    \mathcal{F}(\mathcal{D},\theta) = KL[q(\mathbf{w} | \theta)~||~P(\mathbf{w})] - \mathbb{E}_{q(\mathbf{w} | \theta)}[\log P(\mathcal{D} | \mathbf{w})]
\end{equation}

The second term in this cost function is the negative log likelihood loss which depends on the dataset $\mathcal{D}$. On the other hand the first part depends on the selected prior $P(\mathbf{w})$ for the parameters. Following the strategies of previous works in Bayesian deep learning, we choose a simple prior $\mathcal{N}(0, I)$. 
Since exact calculation of KL divergence is also intractable, Equation \ref{eq:cost} is approximated by drawing $n$ Monte Carlo samples using the following equation:

\begin{equation}
    \label{eq:cost_approx}
    \mathcal{F}(\mathcal{D},\theta) \approx \sum_{k=1}^{n} \log q(\mathbf{w}^{(k)} | \theta) - \log P(\mathbf{w}^{(k)}) - \log P(\mathcal{D} | \mathbf{w}^{(k)})
\end{equation}

Here, $\mathbf{w}^{k}$ is the $k^{th}$ Monte Carlo sample drawn from the posterior estimate $q(w | \theta)$. Note that, cases where closed forms of KL can be computed, Equation \ref{eq:cost_approx} still performs equally well~\cite{blundell2015weight}.

\subsection{Inference}
In the inference step, we calculate both the output probability and aleatoric uncertainty of the output decision (resulting from input noise). The output probability $p^{\prime}$ and the aleatoric uncertainty $\Ddot{u}_{al}$ can be estimated as follows \cite{kwon2018uncertainty}:

\begin{gather}
    p^{\prime} = \frac{1}{n} \sum_{k=1}^{n} \hat{{p}_k}\\
    \Ddot{u}_{al} = \frac{1}{n} \sum_{k=1}^{n} [diag(\hat{{p}_k}) - \hat{{p}_k}\hat{{p}_k}^{T}]
\end{gather}

Here, $\hat{{p}_k} = softmax[f_{w_k}(x)]$ where $w_k$ is the $k$th Monte Carlo sample weight drawn from the posteriors and $x$ is the input signal.

\subsection{Why Bayesian Deep Learning in AF Detection?}
{While we want to detect signal irregularities originating from Atrial Fibrillation, we do not want our model to be misled from signal noise originating from micro-motions. This conflicting objective makes it extremely difficult to differentiate whether irregularities in the signals are generated from signal noise or from possible Atrial Fibrillation. Traditional filtering techniques are not sufficient to eliminate noise in such a scenario. Consequently, it can be very useful if a model can "learn" to handle signal noise while optimizing the original objective function of AF detection. Conventional prediction models lack the representation power to extract features from raw PPG data that express the intricate relationship between different attributes.  Regular deep learning models while having rich representation power, only learn the point estimate of the network parameters. Bayesian Deep learning models on the other hand learn full probability distribution of the parameters while also maintaining the representation power. Thus, during the training of the original objective function, the network parameters in Bayesian Deep learning have the capacity to capture signal noise which makes them robust to perturbations originating from it. These benefits of Bayesian Deep learning makes it highly favorable for AF detection from noisy PPG signals. Furthermore, these models also give uncertainty estimates of model decisions for reliable decision making in real world scenarios.}

\subsection{BayesBeat Model Architecture}

\begin{figure}[t!]
\begin{center}
\includegraphics[width=\textwidth]{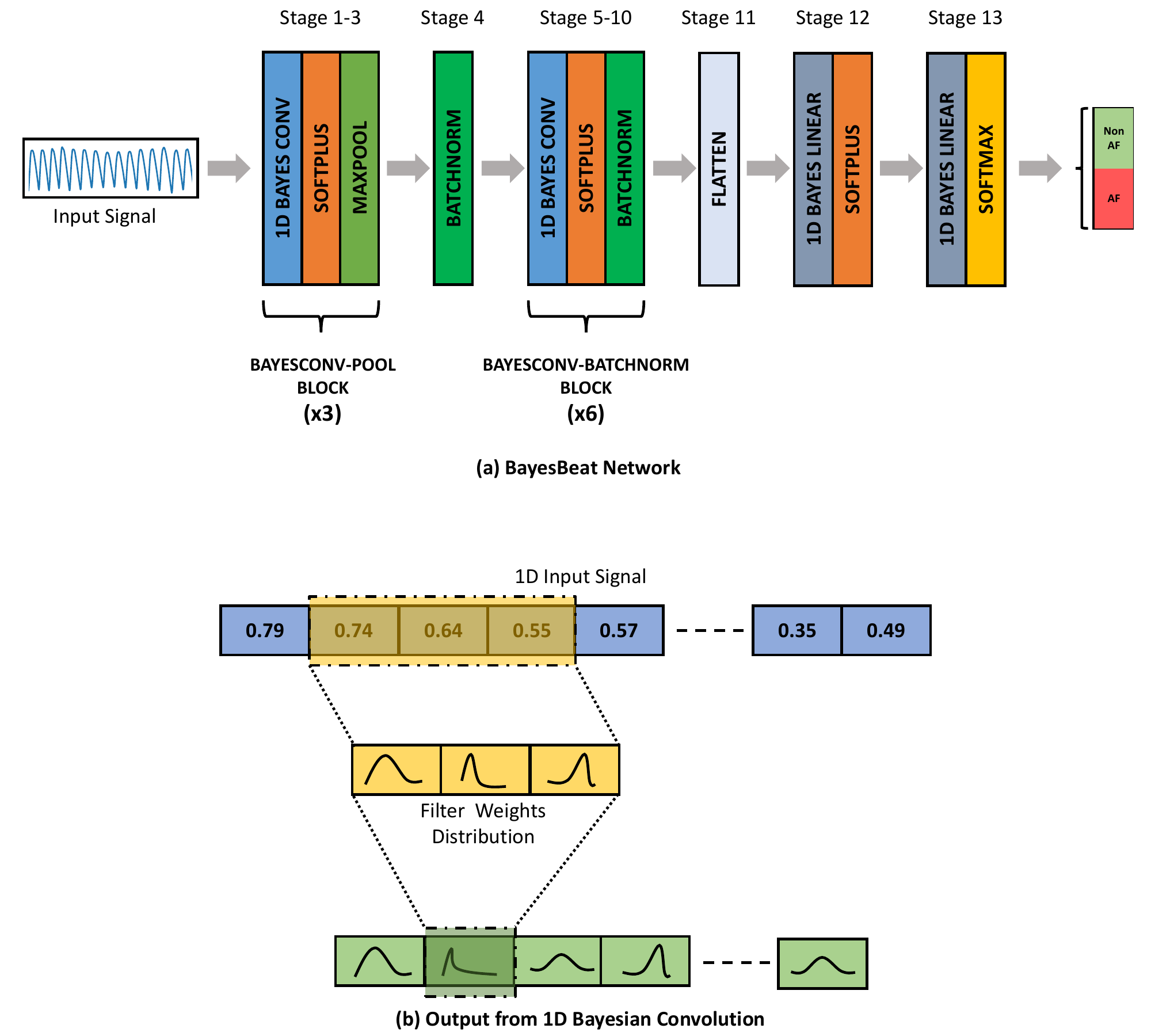}
\caption{ (a) Overview of our proposed \textit{BayesBeat} architecture. (b) Illustration of a 1D Bayesian convolutional layer (BayesConv1D) of filter size 3. Here each weight in the convolution filter corresponds to a posterior distribution instead of a specific value. }

\label{fig:fig_archi}
\end{center}
\end{figure}

Figure \ref{fig:fig_archi}(a) demonstrates the network architecture of \textit{BayesBeat}. The network is composed of a total of nine Bayesian convolutional layers followed by two fully connected layers.
The network takes 1D signal segments from the dataset as input. Then the input is passed to three consecutive blocks, we named them as \textbf{BAYESCONV-POOL} blocks, which transform the input signals into multi-channel 1D features. Each of these \textbf{BAYESCONV-POOL} blocks is composed of Bayesian 1D Convolution with padding to keep the output size of each channel equal to the input and softplus activation followed by a maxpool operation that help us to downsample the input signal removing redundant information and summarizing the key features. The output from the last \textbf{BAYESCONV-POOL} block is then batch-normalized with 1D batch normalization. After that, a sequence of six \textbf{BAYESCONV-BATCHNORM} blocks are executed on this batch-normalized output. Every \textbf{BAYESCONV-BATCHNORM} block has a Bayesian 1D convolution with padding and activation as softplus followed by a 1D batch normalization. \textbf{BAYESCONV-BATCHNORM} blocks do not contain any pooling operation since downsampling the input too much may result in losing key features. These blocks transform the features into multi-channel 1D shrinked features. After these nine bayesian convolutional layers, the features are flattened into a single feature vector. The feature vector is then passed into a bayesian linear layer with softplus activation followed by another bayesian linear layer with softmax activation since it is the final classification layer. Finally, we get the AF/Non-AF prediction as a probability distribution over two classes. At each stage, we employ \emph{softplus} as our preferred activation function as it outperforms rectified linear unit (ReLU) in our case.\\
Note that the bayesian convolutional layer has filters and executes sliding window based convolution operation similar to a traditional deep learning convolution. However as described earlier, unlike the filters of traditional deep learning convolutional layer, every single weight of the filters in bayesian convolution is a posterior distribution instead of a single point value as shown in Figure \ref{fig:fig_archi}(b). We sample layer activations utilizing the local reparameterization trick (LRT) \cite{kingma2015variational} as per the formulation from \cite{shridhar2019comprehensive} for computational acceleration.
Since our inputs are of single dimension, the 1D bayesian convolution has been applied in our case.


In Bayesian deep learning models, batch normalization and dropout can perturb the actual likelihood function, which may affect the overall performance. We face similar outcomes with dropout as \cite{wenzel2020good}, i.e. incorporating dropout decreases model performance.
However, batch normalization induces performance improvement. Consequently, batch normalization is employed for the final six stages of convolution while no dropout is employed at any stage of the network. 

\input{files/pseudocode}

\subsection{Training}
The cost function in Equation \ref{eq:cost} needs to be optimized by \emph{minibatch gradient descent}. Since optimization has to be done on different minibatches of same size, the prior dependent part of Equation \ref{eq:cost} requires to be weighted. Consequently, the cost function for minibatch $i$ becomes:

\begin{equation}
    \label{eq:cost_final}
    \small
    \mathcal{F}_{i}(\mathcal{D}_i,\theta) = \lambda_i KL[q(\mathbf{w} | \theta)~||~P(\mathbf{w})] - \mathbb{E}_{q(\mathbf{w} | \theta)}[\log P(\mathcal{D}_i | \mathbf{w})]
\end{equation}

\subsection{Choosing Proper ${\lambda_{i}}$}
{The choice of $\lambda_i \in [0,1]$ is a crucial factor for the performance of our model. Directly applying $\lambda_i$ value  as suggested by ~\cite{blundell2015weight} resulted in underfitting of the dataset. The  $\lambda_i$ value used in \cite{sonderby2016ladder} also gave us similar problem. On the other hand, $\lambda_i = 1$ gave us convergence problems whereas $\lambda_i = 0$ resulted in severe overfitting. So, as we put more weight on optimizing the KL-divergence from the prior, the model refuses to fit the training dataset. This behavior based on the values of $\lambda_i$ suggests us extremely complex nature of the distribution required for this dataset. A careful choice of $\lambda_i$ is thus vital to the performance of our model.}

{From these observations, we select $\lambda_i$ as a combination of two parameters, $\alpha$ and $\beta$, where $\lambda_i = \alpha\beta$. Here, $\mathbf{\alpha}$ is a scaling term, and in our experiments we have observed that smaller values of $\mathbf{\alpha}$ generally result in better performance and consequently we set it to $10^{-5}$ for optimal performance based on the validation set results. Scaling with such a small weight gives proper regularization effect by encouraging the network about aligning with the prior without interfering with learning the distribution required for the complex training data. On the other hand, $\mathbf{\beta = \frac{2^{M - i}}{2^M - 1}}$, the value used by \cite{blundell2015weight} where $M$ is the total number of minibatches. This $\beta$ term helps to get the advantage of \cite{blundell2015weight} so that the initial training steps get higher regularization where the later steps gets more focused on the training data for an appropriate fit. With this scheme of $\lambda_i$, our network properly converged during training without any overfitting, resulting in good prediction performance.}

\subsection{Implementation Details}

We adapt the training pipeline of \emph{BayesByBackprop(BBB)}~\cite{blundell2015weight} for training our model following the PyTorch~\cite{NEURIPS2019_9015} implementation in \cite{shridhar2019comprehensive}. The details of the training are presented in Algorithm \ref{algo:train_bbb}. While Algorithm \ref{algo:train_bbb} performs reasonably well for training, using local reparameterization trick~\cite{kingma2015variational} on top of that makes training faster. Here, instead of sampling the weights, the activations are sampled, resulting in variance reduction of gradients. We chose $batch\_size = 512$, and trained our network for 50 epochs with a learning rate of $1 \times 10^{-3}$. For optimizer, we chose \emph{Adam}~\cite{kingma2014adam}. During the training, we tuned the hyperparameters and selected the best model based on the validation set performance. The whole model was trained using an i7-7700 workstation with 16 GB of RAM, and a GTX 1070 GPU for 1.5 days. In our experimental study, we have observed that \emph{BayesBeat} converges in much fewer number of epochs compared to both \citt{shen2019ambulatory} and \citt{torres2020multi}.

%% file: files/pseudocode.tex
\begin{algorithm}[htb]
\footnotesize
\DontPrintSemicolon
  
  \KwIn{parameters, $\theta = (\mu, \rho)$, minibatches, $\mathcal{D} = \{\mathcal{D}_1, \mathcal{D}_2, \ldots, \mathcal{D}_M\}$ number of draws, $n$}
  \KwOut{Updated model parameters, $\theta_{out}$ }
  
  \For{i = $1 \to M$}
    {
     $cost = 0$\\
     $\lambda_i = get\_lambda(i, M)$\\
     \For{k = $1 \to n$}
     {
        Sample $\epsilon \sim \mathcal{N}(0,I)$\\
        $w = \mu + \log [1 + exp(\rho)] \circ \epsilon$\\
        $term_{prior} = \log q(\mathbf{w}^{(k)} | \theta) - \log P(\mathbf{w}^{(k)})$ \hfill (Eq. \ref{eq:cost_approx})\\
        $term_{likelihood} = -\log P(\mathcal{D} | \mathbf{w}^{(k)})$\\
        $cost = cost + \lambda_i term_{prior} + term_{likelihood}$
     
     }
    find the gradients by backpropagating $cost$
    with respect to $\theta$\\
    update parameters $\theta = (\mu, \rho)$ with the computed gradients\\
    }
    $\theta_{out} = \theta$
 
\caption{Training \emph{BayesBeat} with \emph{BBB}}
\label{algo:train_bbb}
\end{algorithm}

%% file: files/results.tex
To test the efficacy of our proposed model \textit{BayesBeat}, we extensively evaluate the performance of our approach and compare it with two state-of-the-art deep learning based approaches (\citt{shen2019ambulatory} and \citt{torres2020multi}) for AF detection. 

\subsection{Dataset}
To evaluate different models, we have used the largest publicly available dataset, which we refer to as the \textit{Stanford Wearable Photoplethysmography Dataset}\footnote{https://www.synapse.org/\#!Synapse:syn21985690/files/}. This dataset was recently made public by~\citt{torres2020multi} with their work \emph{DeepBeat}. The dataset contains more than 500K signal segments from a total of {175 individuals (108  AF subjects and 67 non-AF subjects)} collected using wrist-worn wearable devices~\citt{torres2020multi}. Each of the data segments is 25s long sampled at 128 Hz and later downsampled to 32 Hz. The starting timestamp of each segment is also provided along with the dataset. This dataset contains sufficiently noisy signal segments that conform to the real-world settings in the ambulatory environment.

The dataset provided is already preprocessed, that is all the signals are bandpass-filtered, and standardized to [0,1]. Hence, no further preprocessing has been carried out.
The dataset is also supplemented with three types of signal labels, i.e., Poor, Good, and Excellent. However, only a small portion of them are labeled by human; the rest of them are generated from a model trained on the human-labeled portion. While this process makes it easy to label such a huge dataset, it also generates some imprecise labels in the dataset which can make downstream models vulnerable to error propagation. Figure \ref{fig:fig_comparesignals} shows such instances in red color where the accompanying quality labels do not represent the true quality of the signal. Considering these pitfalls, instead of using the given labels, we use the uncertainty estimation from our BayesBeat model to account for the prediction.

\input{files/dataset_dist_table}

\noindent \textbf{Data Splitting:} First, we address some distribution issues present in the original train, validation, test split of the dataset. In the originally published dataset, the test set is extremely small and there are overlapping signal segments that introduce repetition of the same segments in test samples. Furthermore, the ratio of AF and non-AF subjects are also quite different from that of the train and validation sets. Table \ref{tab:dataset_table_deepbeat} presents a description of the dataset which highlights the issues in the original split of the dataset. Due to the above issues, the test set can misrepresent the performance of a model. We find that while the proposed model in~\citt{torres2020multi} reported exceptional performance in test set (sensitivity: 0.98, specificity: 0.99, F1-score: 0.96), the performance is not satisfactory in the validation set (sensitivity: 0.59, specificity: 0.995, F1 score: 0.69) and train set (sensitivity: 0.59, specificity: 0.998, F1 score: 0.74). 

{To address these problems, we redistribute the samples in the dataset and create a split of 70\% as train, 15\% as validation, and 15\% as test sets. During the redistribution of the dataset, we reshuffled subjects among the sets and also ensured that no subject is shared among different sets}. {Furthermore, we removed all overlapping signal segments from the validation and test sets using the provided recording timestamp of each segment to ensure that the same windows are not counted multiple times during evaluation. However, overlapping in train set has been retained since this serves the purpose of data augmentation. Moreover, no subjects were shared across train, validation, and test sets. In our split, the ratio of AF and non-AF segments in test set (0.39) is also very close to the actual dataset (0.38) which ensures that we get an accurate representation of the entire dataset. Table \ref{tab:dataset_table_bayesbeat} provides a description of the our split. This redistributed form of the dataset will be available in our source code repository.
}

\subsection{Evaluation Metrics and Baselines}
To evaluate different models, we have used a wide range of metrics that include Recall (Sensitivity), Specificity (True Negative Rate, TNR), Precision (Positive Predictive Value, PPV), and F1-score. These metrics can be formally expressed as follows.
\begin{small}
\vspace{-7mm}
\begin{multicols}{2}
\begin{equation*}
    Recall = \frac{TP}{TP + FN}
\end{equation*}

\begin{equation*}
    Specificity = \frac{TN}{TN + FP}
\end{equation*}

\begin{equation*}
    Precision = \frac{TP}{TP + FP}
\end{equation*}

\begin{equation*}
    F1 = \frac{2*(Precision*Recall)}{Precision + Recall}
\end{equation*}

\end{multicols}
\end{small}

Here, TP (True Positive) is the number of positive signals correctly classified by the model, TN (True Negative) is the number of negative signals correctly classified by the model, FP (False Positive) is the number of signals falsely classified as positive, and FN (False Negative) is the number of signals falsely classified as negative.

We also use AUC, area under ROC curve and \textit{Matthews Correlation Coefficient (MCC)} as our evaluation metrics. Though AUC is commonly used in the prior works, it can give misleading performance representation for class imbalanced dataset; whereas MCC is particularly effective in assessing the performance of a binary classifier in case of imbalanced data~\cite{boughorbel2017optimal}. MCC is defined as follows:

\begin{small}
\begin{equation*}
        MCC = \frac{TP*TN - FP*FN}{\sqrt{(TP+FP)(TP+FN)(TN+FP)(TN+FN)}}
\end{equation*}
\end{small}

We compare our results against the most recent state-of-the-art works by~\citt{shen2019ambulatory} and ~\citt{torres2020multi}. Since they demonstrated that their methods significantly outperform traditional statistical and classical machine learning base line models, we focus on comparing with these state-of-the-art deep learning based approaches. For~\citt{shen2019ambulatory}, we implemented the 1D ResNeXt architecture having over 7.5 million parameters with the same network and hyperparameter settings as stated in the paper. Since the network  does not explicitly take into account noise in the data, we found that it fits the training data quite easily irrespective of signal quality, which eventually leads to overfitting. On the other hand, \citt{torres2020multi} provided the architecture of their model ``\emph{DeepBeat}'' which we first pre-train as in and then train on the redistributed dataset. This baseline has a separate signal quality estimation part, with which the excellent quality signals can be separated first, and then the model performance can be measured only on the excellent signals. Therefore, we include its performance both in the ``raw'' form and in excellent signals only (denoted as ``Excellent only''). {Since, a large percentage of samples are discarded if we consider only excellent signals, we have also calculated the performance metrics excluding poor signals only i.e. with excellent and good signals.}

\section{Results}

In this section, we present the results of our experiments. First, we analyze the performance of Bayesbeat for various uncertainty thresholds and then we compare it with state-of-the-art methods.

\subsection{Model Performance without Uncertainty Threshold}

{Table~\ref{tab:Perf_table_bayesbeat} presents the performance summary of \emph{BayesBeat} on the redistributed dataset
without any uncertainty threshold i.e. on the entire dataset. We observe that  the performance metrics are slighly lower in validation and test sets compared to the training set, which is due to data augmentation in the training set in addition to possible overfitting. However, values of metrics are consistent between validation and test set, unlike the performance of Deepbeat on the data split provided by~\citt{torres2020multi}}

\input{files/performance_tab_bayesbeat}

\input{files/analysis}

\input{files/performance_test}

\subsection{Comparison with State-of-the-art Methods}

Table~\ref{tab:Perf_table_test} shows the performances of different models, where we compare sensitivity (recall), specificity, precision, F1-score, AUC, MCC and coverage percentage of \textit{BayesBeat} against state-of-the-art methods on test set of the redistributed dataset.
{We show performances of \emph{BayesBeat} for no uncertainty threshold i.e., on the entire test set as well as for uncertainty bound of 0.05. The two sets of metrics essentially demonstrate how \emph{BayesBeat} generates high uncertainty scores for noisy signals and filtering such signals by setting low uncertainty threshold substantially improves detection performance without resorting to any explicit signal quality estimation training that might itself suffer from erroneous quality labelling.}


The table also shows performance metrics of \emph{DeepBeat}~\citt{torres2020multi} on three settings as well as those of ~\citt{shen2019ambulatory}. Additionally, we also include the performances of simple DNN baseline paired with wavelet denoising \cite{lee2003reduction} and DTCWT \cite{raghuram2012dual}. We observe that \textit{BayesBeat}, with its uncertainty estimation capability, substantially outperforms all the state-of-the-art methods.
Compared to~\citt{shen2019ambulatory}, \textit{BayesBeat} achieves 7\% higher F1 score (when uncertainty threshold of 0.05 is used). BayesBeat also outperforms \emph{DeepBeat}~\citt{torres2020multi}) by large margins (10-20\% depending on the setting of \emph{DeepBeat}). 

We note that while \emph{DeepBeat} achieves a sensitivity score of 0.8 when applied to signals of all qualities, this severely affects other metrics resulting in the lowest performance in terms of AUC and MCC.
While \emph{DeepBeat} achieves relatively better metrics when only excellent signals are used, it covers less than 19\% of the signals in the test set in that case. 
Again, even if we consider \emph{DeepBeat} with that only signals predicted to be excellent, \textit{BayesBeat} with uncertainty threshold of 0.05 still outperforms \emph{DeepBeat} with significant margins with a coverage of over 54\%.

It is worth highlighting that although~\citt{shen2019ambulatory} outperforms \textit{BayesBeat} and \emph{DeepBeat} when applied to the full dataset, it cannot filter out noisy signal segments which are unfit for analysis according to~\cite{elgendi2016optimal}. Therefore, its performance would deteriorate in conditions where large portions of the signals could be noisy and hence is unsuitable for ambulatory settings. 

\begin{figure}[!htb]
\begin{center}
\includegraphics[width=0.55\textwidth,height=0.35\textheight]{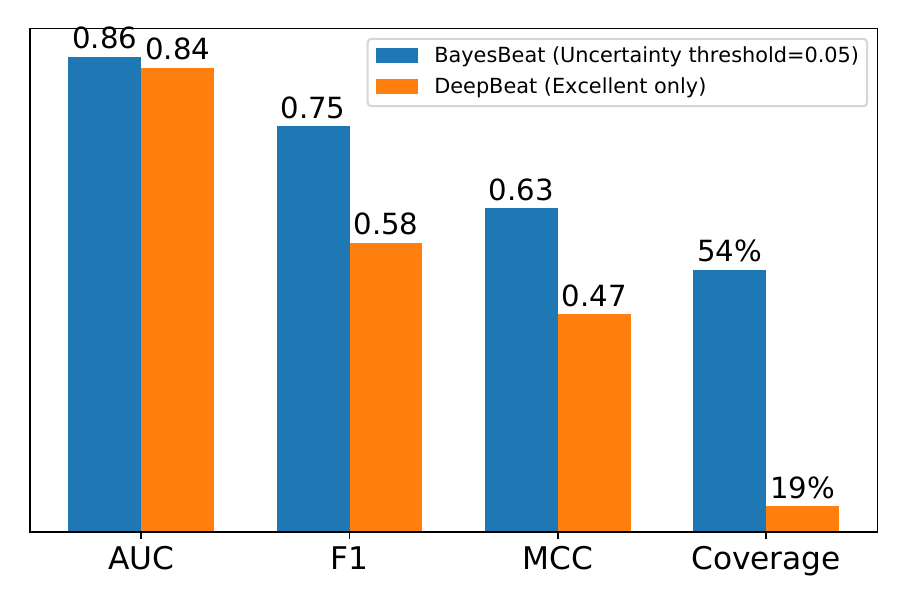}
\caption{Performance metrics and test set coverage comparison between \textit{BayesBeat} (uncertainty threshold=0.05) and \emph{DeepBeat} (excellent only). \textit{BayesBeat} (threshold=0.05) has significantly larger coverage than the \emph{DeepBeat} (excellent)}
\label{fig:bayesbeat_deepbeat_compare}
\end{center}
\end{figure}

Finally, we observe that simple DNN model paired with sophisticated signal processing techniques performs poorly compared to \emph{BayesBeat} and all the other baselines. This indicates that even sophisticated signal processing techniques are not sufficient to handle the noise present in ambulatory PPG signals. On the other hand, both \textit{BayesBeat} and \emph{DeepBeat} try to filter out signals that are unfit for analysis. Figure~\ref{fig:bayesbeat_deepbeat_compare} shows a direct comparison between \textit{BayesBeat} with bounded uncertainty score of 0.05 and \emph{DeepBeat} with only excellent signals. In summary, \emph{BayesBeat} outperforms the state-of-the-art methods, and at the same time provides an uncertainty estimate of the results based on the signal noise, which indicates the efficacy and robustness of our proposed approach.


\subsection{Transformed Feature Map Visualization}

\begin{figure}[hp]
    \vspace{-3mm}
    \begin{subfigure}[h]{\textwidth}
        \centering
        \hspace*{-12.25mm}
        \includegraphics[height=0.32\textheight, width=0.9\textwidth]{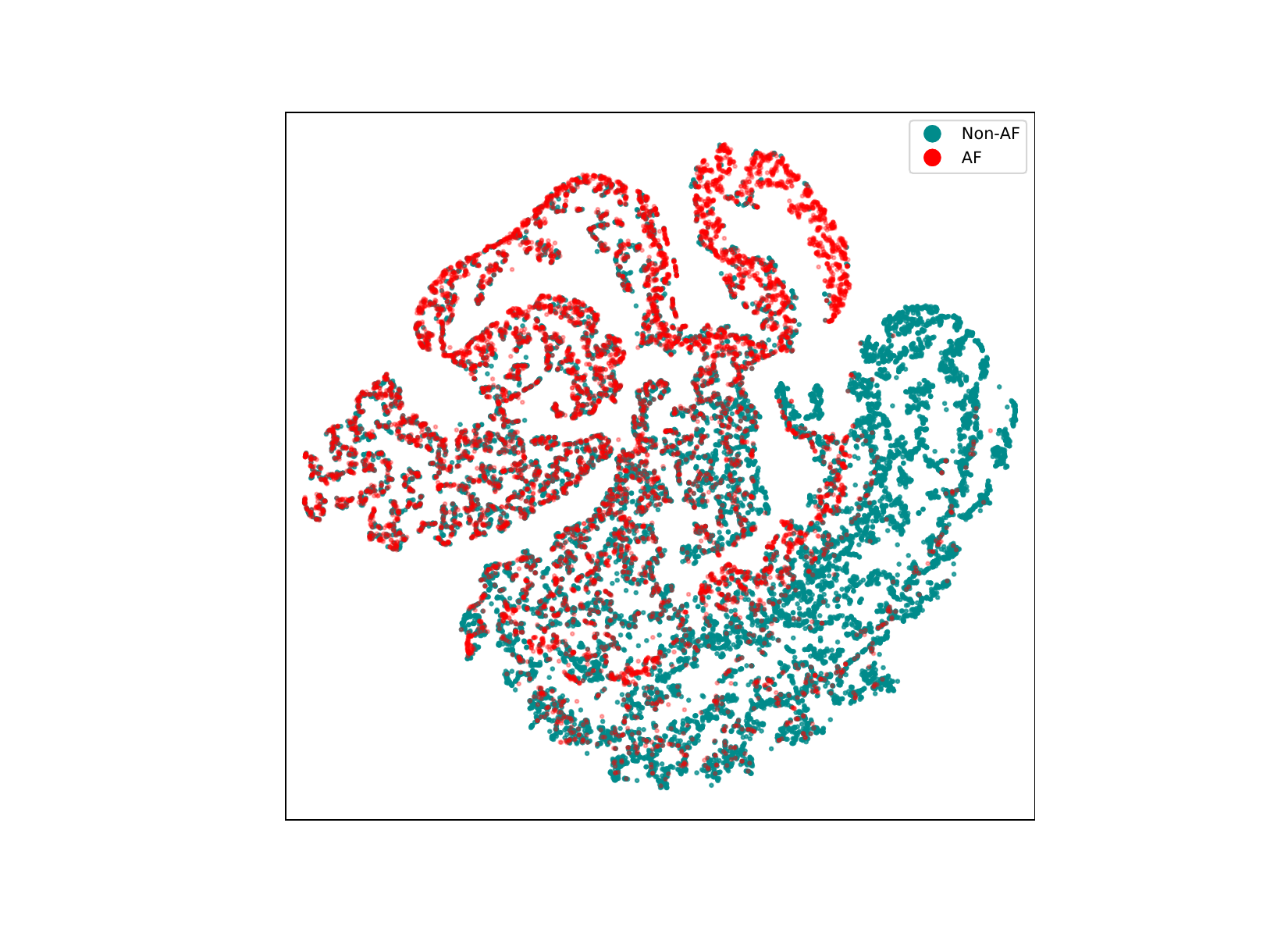}
        \vspace{-2.25\baselineskip}
        \caption{Without any bound on Uncertainty score}
        \label{fig:tsne:all}
    \end{subfigure}
    \begin{subfigure}[h]{\textwidth}
        \centering
        \includegraphics[height=0.34\textheight, width=0.9\textwidth]{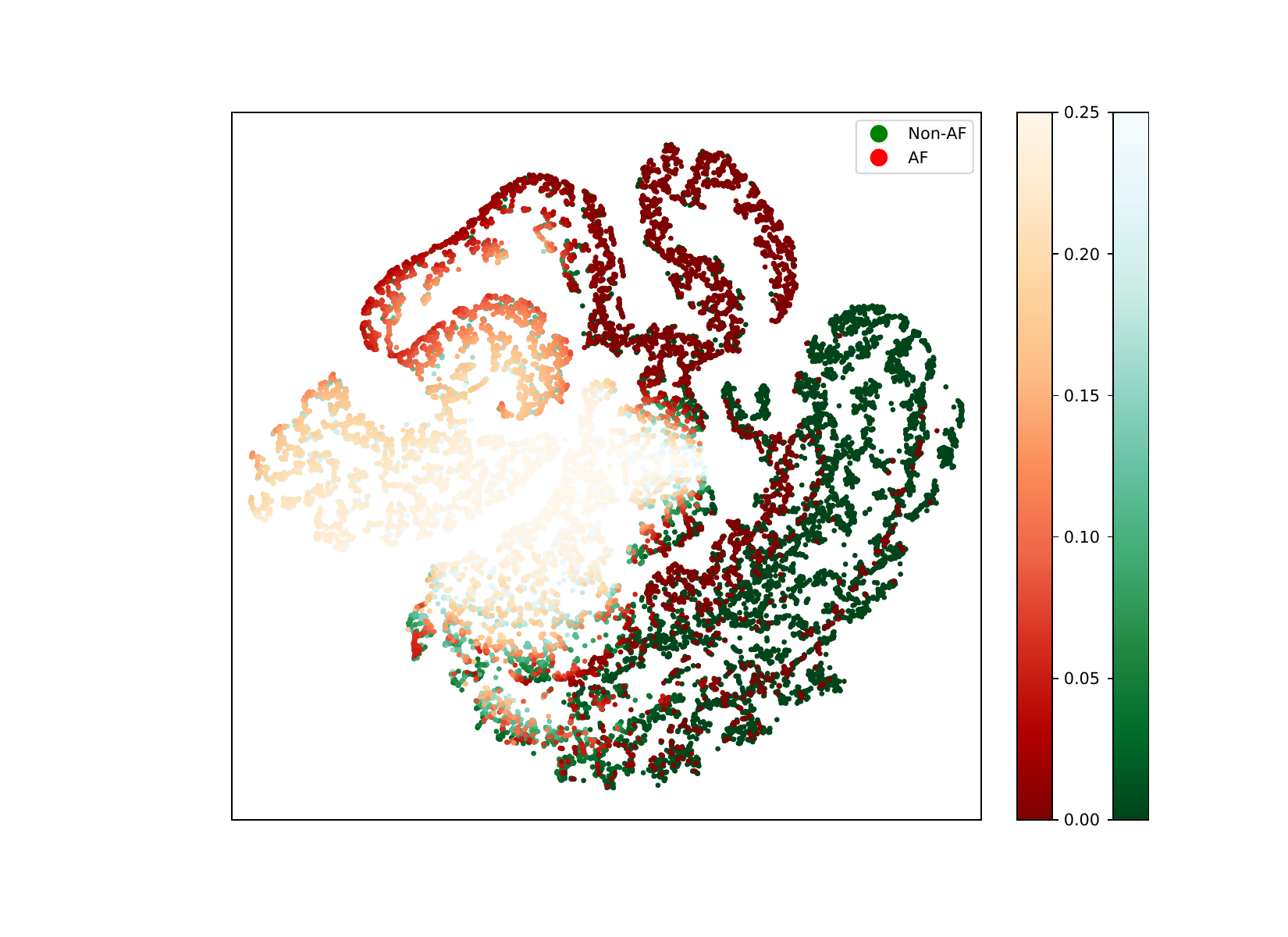}
        \vspace{-2.25\baselineskip}
        \caption{Uncertainty score intensity for each point. Light color represents higher uncertainty score}
        \label{fig:tsne:intensity}
    \end{subfigure}
    \begin{subfigure}[h]{\textwidth}
        \centering
        \hspace*{-12.25mm}
        \includegraphics[height=0.34\textheight, width=0.9\textwidth]{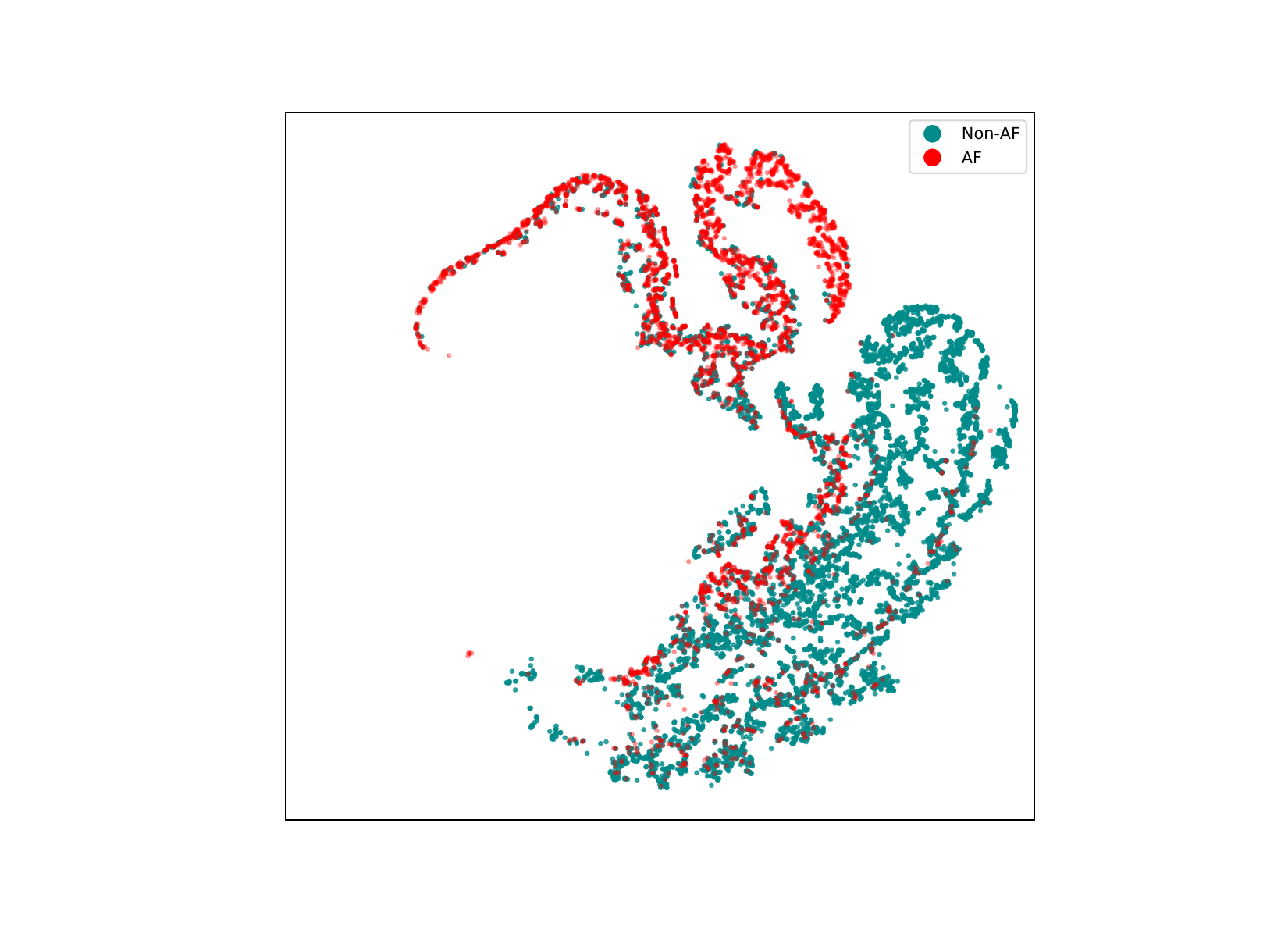}
        \vspace{-2.25\baselineskip}
        \caption{After setting the Uncertainty bound to 0.05}
        \label{fig:tsne:bounded}
    \end{subfigure}
\caption{t-SNE visualization of the transformed feature map from the linear layer of our network. A prominent separation between the Non-AF and AF records in this feature space demonstrates the potency of \emph{BayesBeat}}

\label{fig:tsne}
\end{figure}

Figure \ref{fig:tsne} shows the t-SNE visualizations~\cite{maaten2008visualizing} of the transformed features from the linear layer of our network over different settings on test set. We have presented three t-SNE figures for three different settings. The first plot in Figure~\ref{fig:tsne:all} contains all the points from the test set without applying any uncertainty score based filtering. Therefore, we can observe that there are more mixture of AF(Cyan) and Non-AF(Red) points.\\
Figure~\ref{fig:tsne:intensity} represents a corresponding visualization of Figure~\ref{fig:tsne:all} which contains every points of the first one but every point receives color intensity (dark or light) based on their uncertainty score. The light colored points are with higher uncertainty scores, light red and light green are for AF and Non-AF points, respectively, whereas uncertainty scores are lower for dark colored points. As we can see that the middle portion that has the most mixture of AF and Non-AF points in Figure~\ref{fig:tsne:all} received mostly the light color i.e. has higher uncertainty scores among other points in the figure. These are the points corresponding to noisy segments and so removing them will increase the performance.\\
In Figure~\ref{fig:tsne:bounded}, we filter out the points from Figure~\ref{fig:tsne:all} those received higher uncertainty score than 0.05 to demonstrate the impact of filtering on this test set with uncertainty threshold of 0.05. We can see the separation between the AF and non-AF records is quite prominent in this transformed feature space generated by our model. Most of the points that received light colors and had more mixture are filtered out in Figure~\ref{fig:tsne:bounded}. However, there are some minor presence of AF cluster in Non-AF cluster. These are the records for which the network is relatively less confident about whether the rhythm irregularity is due to Atrial Fibrillation or from the input noise. This demonstrates that \emph{BayesBeat} effectively learns to distinguish between AF and non-AF PPG records even in challenging situations.

\subsection{Performance Comparison on MIMIC-III Waveform Database Matched Subset}

In this section, we present a performance comparison of Bayesbeat with state-of-the-arts on a AF labeled subset of the Medical Information Mart for Intensive Care (MIMIC) III dataset, a large open source medical record database that is publicly available in PhysioNet~\cite{goldberger2000physiobank}. The database contains simultaneous recordings of both ECG and PPG signals for individuals. However, physionet does not provide AF labels for the signal segments.
Consequently, we get the labels from ~\cite{bashar2020atrial}, which contains the human annotations for ECG signals parallel to the PPG recordings for 35 individuals in MIMIC-III. Among them, 19 individuals had AF where the rest 16 individuals were non-AF. This results in a total of 125629 segments of PPG signals, on which we evaluate BayesBeat against state-of-the-art methods. We show these results in Table~\ref{tab:mimic3_perf_comp}. Note that Bayesbeat and all the state-of-the-art methods we compare in this table are trained on redistributed DeepBeat dataset.

\input{files/performance_mimic3}

Table~\ref{tab:mimic3_perf_comp} represents that \emph{Bayesbeat} outperforms all the state-of-the-art methods even on MIMIC-III dataset. Note that all these models have been trained on Deepbeat dataset. This signifies the robustness and generalizability of Bayesbeat over different datasets without requiring any further dataset specific tuning. Note that, although there is no signal quality label available for this dataset, these signals were acquired in ICU and hence they are of better quality than the Deepbeat dataset. This fact is even represented by the significantly higher coverage percentage filtered with the same uncertainty threshold than of the Deepbeat dataset. Another fact from this table is that, Bayesbeat achieves the highest F1 score, 13.6\% higher than~\citt{shen2019ambulatory} and 10.2\% higher than Deepbeat~\citt{torres2020multi} even without filtering out any signals by uncertainty threshold. It is also visible from the table that Bayesbeat with any uncertainty bound reported in the table achieves both better performance and significantly higher coverage than signal quality label based filtering method (Non-poor or Excellent only) of Deepbeat. \\
We have reported performances of Bayesbeat for three separate uncertainty thresholds for this dataset in Table~\ref{tab:mimic3_perf_comp} to demonstrate the fact that performance gradually increases with stricter uncertainty bounds at a cost of coverage percentage. Thus, it implies that the uncertainty threshold is a setting that can be selected based on the environment to find a trade-off between performance \& coverage.

\subsection{Efficiency and Flexibility}
Besides state-of-the-art performance results, our method is significantly more memory efficient compared to other approaches. Training Bayesbeat requires only 180K parameters (750KB) in total, which is significantly more lightweight than~\citt{torres2020multi} \textit{DeepBeat} (40 million parameters/155MB) and \citt{shen2019ambulatory} (7.5 million parameters/27MB). We also calculate the inference time of the models in a Google Colab VM with an Intel(R) Xeon(R) CPU @ 2.20GHz with only two available cores. We find that on average AF detection requires 6.58ms which is significantly faster than the 28.09ms for \citt{shen2019ambulatory} and slightly slower than poorly performing DeepBeat \citt{torres2020multi} (4.20ms) that does not utilize all its parameters. This makes Bayesbeat highly suitable for deployment in mobile and wearable devices with very limited memory capacity and computational power. Bayesbeat also gives us an additional dial to tune the trade-off between accuracy and uncertainty, that makes it useful for deployment in commercial smartwatches. Using this dial, smartwatch manufacturers can flexibly tune the deployed model according to the user needs

%% file: files/dataset_dist_table.tex

\begin{table*}[hbt!]
\begin{center}
\caption{\label{tab:dataset_table_deepbeat} Distribution of the Stanford Wearable Photoplethysmography Dataset}

\begin{tabulary}{\linewidth}{|C|C|C|c|C|C|C|C|C|C|C|C|} 
    \hline
    
    Set & \#Individuals & \#Samples (incl. overlapping) & \#AF Samples & \#Non-AF Samples & \#Poor (Labelled) Samples & \#Good (Labelled) Samples & \#Excellent (Labelled) Samples & Data ratio & AF ratio \\
    \hline
    Train & 137 & 2799784 & {1269660} & 1530124 & 1968058 & 281024 & 550702 & 0.839 & 0.45 \\
    \hline
    Validation & 16 & 518782 & 47607 & 471175 & 329140 & 64647 & 124995 & 0.156 & 0.09 \\
    \hline
    Test & 22 & 17617 & 4230 & 13387 & 12339 & 2032 & 3246 & 0.005 & 0.24 \\
    \hline
    
\end{tabulary}
\end{center}
\end{table*}

\begin{table*}[hbt!]
\begin{center}
\caption{\label{tab:dataset_table_bayesbeat} Distribution of the Revised Dataset for Bayesbeat}
\begin{tabulary}{\linewidth}{|C|C|c|C|C|C|C|C|C|C|C|C|C}     
    \hline
    Set & \#Individuals & \#Samples & \#AF Samples & \#Non-AF Samples & \#Poor Samples & \#Good Samples & \#Excellent Samples & Data ratio & AF ratio & \#Segments (incl. overlapping) \\
    \hline
    Train & 132 & {108171} & 41511 & 66660 & 74926 & 11822 & 21423 & 0.698 & 0.38 & 2238821 \\
    \hline
    Validation & 20 & 22294 & 8310 & 13984 & 15858 & 2139 & 4297 & 0.144 & 0.37 & 22294 \\
    \hline
    Test & 23 & 24579 & 9703 & 14876 & 17219 & 2386 & 4974 & 0.159 & 0.39 & 24579 \\
    \hline

\end{tabulary}
\end{center}
\end{table*}

%% file: files/performance_tab_bayesbeat.tex
\begin{table}[h]
\begin{center}
\caption{\label{tab:Perf_table_bayesbeat}Performance of Bayesbeat without uncertainty threshold}
\begin{tabular}{lcccccc}
    \toprule
     & Sensitivity & Specificity & Precision & F1 & AUC & MCC \\
    \midrule
    Train & 0.905 & 0.802 & 0.765 & 0.829 & 0.945 & 0.697 \\
    Validation & 0.678 & 0.71 & 0.581 & 0.626 & 0.754 & 0.379 \\
    Test & 0.722 & 0.72 & 0.627 & 0.671 & 0.793 & 0.433 \\
    \bottomrule
\end{tabular}
\end{center}
\end{table}

%% file: files/analysis.tex
\subsection{Uncertainty Estimation}

\begin{figure}[h]
\begin{center}
\includegraphics[width=\textwidth]{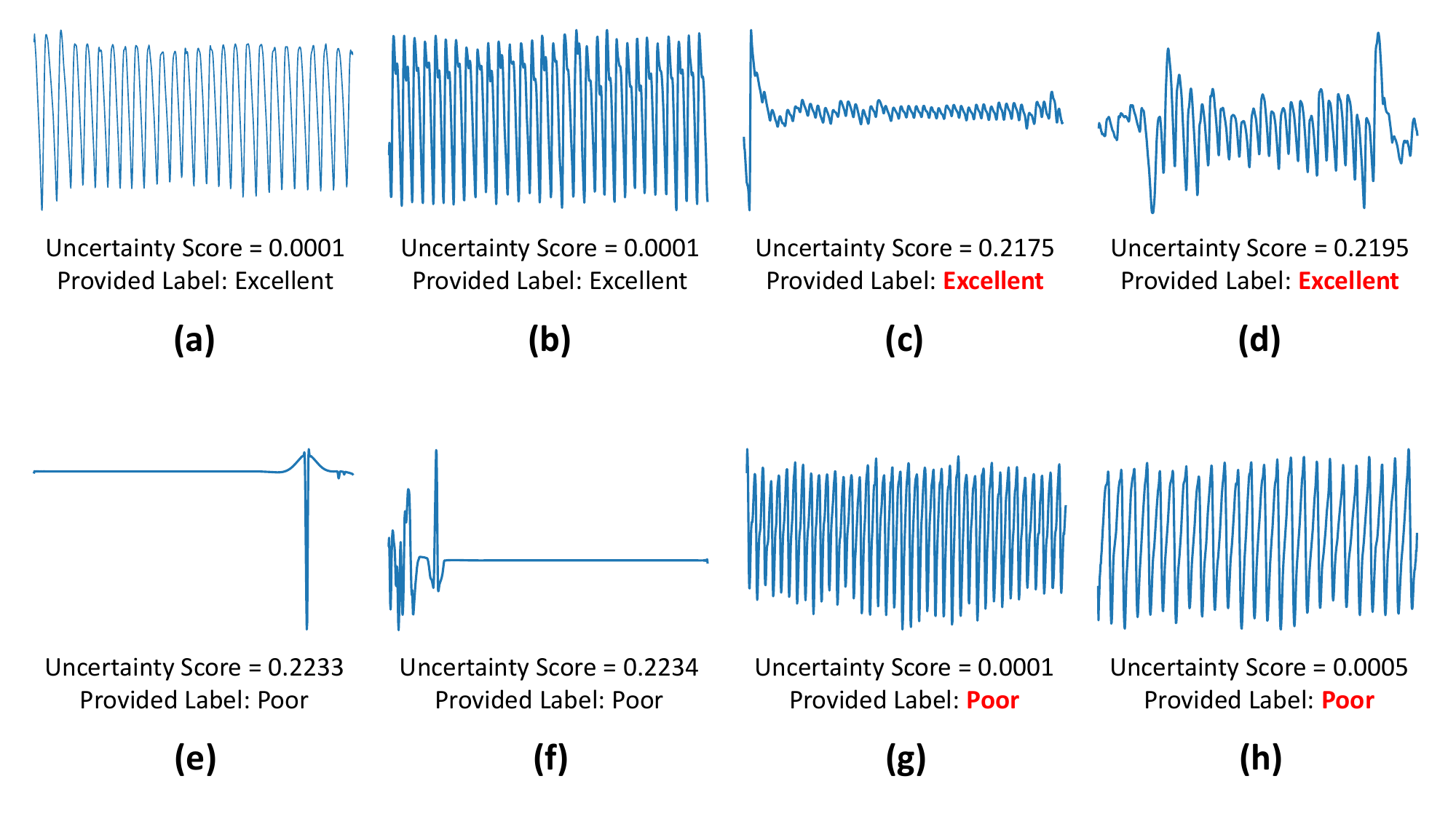}
\caption{Uncertainty scores of different types of signals from our test set. \textbf{Signals (a) and (b)} has minimal degree of noise and \emph{BayesBeat} shows very low uncertainty score; and the provided dataset also labels it as Excellent. \textbf{Signals (c) and (d)} on the other hand is quite noisy with very irregular pattern. Thus our model gives high uncertainty scores to these signals, which the provided dataset labels fail to do.  \textbf{Signals (e) and (f)} has almost no useful PPG data in it, hence \emph{BayesBeat} generates very high uncertainty score, matching the provided label of poor. Finally, \textbf{Signals (g) and (h)} again contain minimal noise showing regular pattern. \emph{BayesBeat} thus generates low uncertainty scores for these signals, which the provided dataset mislabels as ``poor".}

\label{fig:uncertain_ex}
\end{center}
\end{figure}

As discussed above, uncertainty estimation is a vital part of our model which makes it robust against PPG signal noise. 
In Figure~\ref{fig:uncertain_ex}, we present a few representative signals from the test set along with the generated aleatoric uncertainty scores from our model using the method from ~\cite{shridhar2018uncertainty} and the label provided in the dataset. We can see that signal (a) and signal (b) are less noisy, and the beat patterns are also clearly visible, which is the key information required for AF detection. Hence, our model generates very low uncertainty scores (around 0.0001 for both) whereas signal (e) and signal (f) in this figure are heavily perturbed from input noise losing almost all input features that can be used for AF detection. In this case, our model generates high uncertainty scores (0.2233 and 0.2234) indicating that it is less certain about the decision from these inputs demonstrating that the utility of uncertainty scores in \emph{BayesBeat}. The uncertainty scores can be used to filter out noisy signals. They can also be used for weighted averaging of signal segments at various time points which may lead to greater robustness.\\
The uncertainty score generation is also quite accurate for the signal segments that have been inaccurately labelled in original the dataset. For example, signal (c) and signal (d) are labelled as excellent. However, it is clearly visible that perturbation from input noise on both of them are quite indicating and hence uncertainty scores for both of these segment are significantly higher (0.2175 and 0.2195 respectively). The similar improper labelling issues have also been addressed by uncertainty score for signal (g) and signal (h). Although both of them have been labelled as poor, they are quite less noisy and have uncertainty scores of 0.0001 and 0.0005 respective generated from our model. Thus, uncertainty scores generated from our model can successfully address the input PPG signal noise issue without any prior presence of quality labelling. Figure~\ref{fig:uncertain_ex} clearly shows us that although the provided labels are inconsistent across the \emph{DeepBeat} dataset, \emph{BayesBeat} is able to generate consistent uncertainty scores concordant to the true signal quality.

\begin{figure}[h]
\begin{center}
\hspace*{15mm}
\includegraphics[width=0.85\textwidth,height=0.45\textheight]{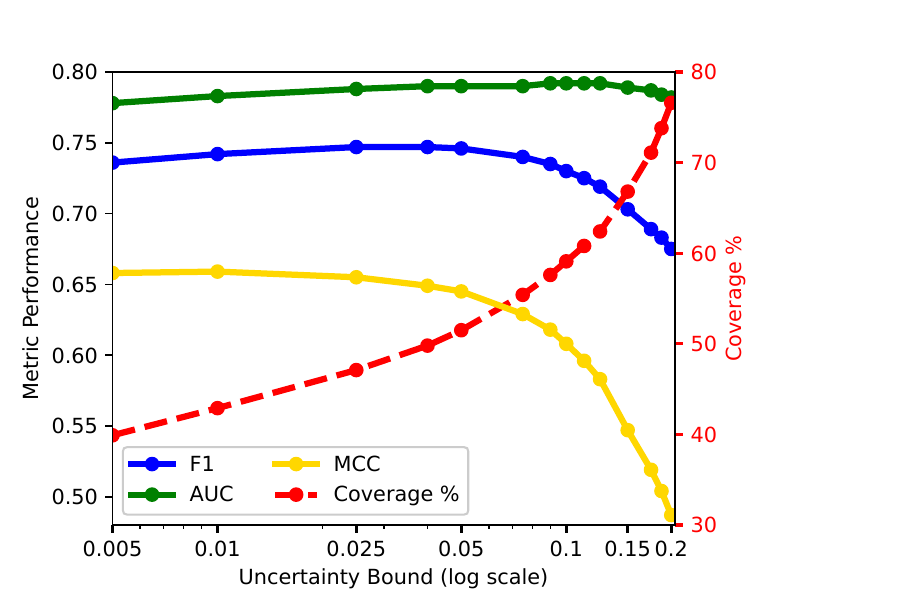}
\caption{F1, AUC \& MCC scores and {Coverage percentage (percentage of signal segments accepted within uncertainty threshold in analysis)} from our validation set for varying uncertainty threshold. A lower uncertainty threshold results in more confident prediction but at a cost of discarding further percentage of signals that have relatively higher uncertainty scores.}
\label{fig:uncertainty_bounds_in_val}
\end{center}
\end{figure}

{Next we analyze the performance of \emph{BayesBeat} for various uncertainty thresholds. 
Figure \ref{fig:uncertainty_bounds_in_val} shows the performance of our model on the validation set for different uncertainty bounds. 
Here, \textbf{coverage refers to the percentage of signals for which the uncertainty score is below the bound and hence considered as valid outputs and used in calculation of performance metrics}.

As we decrease the threshold value, the performance of the model is expected to improve as more signals with high uncertainty scores are discarded. From this figure we can observe that performance on the validation set gets better as we gradually decrease the uncertainty bound until the threshold 0.01. However, below that value, the bound becomes too strict that it starts to reject clean signals resulting in performance degradation. Although we can have better performance with a lower uncertainty threshold, we choose 0.05 as  the prescribed uncertainty bound as it provides a good trade-off between coverage and performance.
}

%% file: files/performance_test.tex
\begin{table*}[hbt]
\begin{center}
\caption{\label{tab:Perf_table_test}Performance of AF detection on the test set from Photoplethysmography data. Lowest performance for each metric is shown in \textcolor{gray}{gray} while the highest performance is \textbf{bolded}}
\scalebox{.92}{
\begin{tabular}{lccccccc} 
    \toprule
     & Sensitivity & Specificity & Precision & F1 & AUC & MCC & Coverage(\%)\\
    \midrule Wavelet Denoising + simple DNN & 0.654	&	\textcolor{gray}{0.505}	&	0.463	&	0.542	&	0.617	&	0.157	&	100\\
    \\
    DTCWT\citx{raghuram2012dual} + simple DNN & \textcolor{gray}{0.585}	&	0.528	&	\textcolor{gray}{0.447}	&	\textcolor{gray}{0.507}	&	\textcolor{gray}{0.586}	&	\textcolor{gray}{0.111}	&	100\\\\
    
    \pbox{20cm}{\citx{shen2019ambulatory}} & 0.664	&	0.819	&	0.706	&	0.684	&	0.806	&	0.489	&	100\\
    \\
    
    \pbox{20cm}{\citx{torres2020multi} \\ (\emph{Deepbeat})} & \textbf{0.803}	&	0.568	&	0.548	&	0.652	&	0.774	&	0.37	&	100\\
    \\
    
    
     
    \pbox{20cm}{\citx{torres2020multi} \\ (\emph{Deepbeat}, Non-poor)} & 0.721	&	0.83	&	0.585	&	0.646	&	0.852	&	0.517	&	27.67\\
    \\
    \pbox{20cm}{\citx{torres2020multi} \\ (\emph{Deepbeat}, Excellent only)} & 0.688	&	0.845	&	0.498	&	0.58	&	0.841	&	0.474	&	\textcolor{gray}{18.77}\\
    \\
    \textbf{BayesBeat} & 0.722	&	0.72	&	0.627	&	0.671	&	0.793	&	0.433	&	100\\
    \\
    \pbox{20cm}{\textbf{BayesBeat} \\ (uncertainty threshold = 0.05)} & 0.728	&	\textbf{0.892}	&	\textbf{0.783}	&	\textbf{0.754}	&	\textbf{0.858}	&	\textbf{0.632}	&	54\\
    \bottomrule
\end{tabular}}
\end{center}
\end{table*}


%% file: files/performance_mimic3.tex
\begin{table*}[hbt]
\begin{center}
\caption{\label{tab:mimic3_perf_comp}Performance of AF detection on the  MIMIC-III Waveform Database Matched Subset. Lowest performance for each metric is shown in \textcolor{gray}{gray} while the highest performance is \textbf{bolded}}
\begin{tabular}{lccccccc} 
    \toprule
     & Sensitivity & Specificity & Precision & F1 & AUC & MCC & Coverage(\%)\\
    \midrule
    \pbox{20cm}{\citx{shen2019ambulatory}} & 0.56 & 0.937 & 0.886 & 0.686 & \textcolor{gray}{0.711} & 0.545 & 100\\
    \\
    
    \pbox{20cm}{\citx{torres2020multi} \\ (\emph{Deepbeat})} & 0.692 & \textcolor{gray}{0.8} & \textcolor{gray}{0.751} & 0.72 & 0.81 & 0.495 & 100\\
    \\
    \pbox{20cm}{\citx{torres2020multi} \\ (\emph{Deepbeat}, Non-poor)} & 0.54	&	0.892	&	0.76	&	0.632	&	0.785	&	0.472	&	61\\
    \\
    \pbox{20cm}{\citx{torres2020multi} \\ (\emph{Deepbeat}, Excellent only)} & \textcolor{gray}{0.282}	&	0.959	&	0.804	&	\textcolor{gray}{0.417}	&	0.773	&	\textcolor{gray}{0.344}	&	\textcolor{gray}{33.8}\\
    \\
    BayesBeat & 0.786	&	0.889	&	0.86	&	0.822	&	0.911	&	0.681	&	100\\
    \\
    \pbox{20cm}{BayesBeat \\ (uncertainty threshold = 0.1)} & 0.804	&	0.944	&	0.924	&	0.86	&	0.925	&	0.762	&	88.1\\
    \\
    \pbox{20cm}{BayesBeat \\ (uncertainty threshold = 0.05)} & 0.81	&	0.953	&	0.933	&	0.867	&	0.928	&	0.778	&	84.6\\
    \\
    \pbox{20cm}{BayesBeat \\ (uncertainty threshold = 0.01)} & \textbf{0.821}	&	\textbf{0.966}	&	\textbf{0.949}	&	\textbf{0.881}	&	\textbf{0.931}	&	\textbf{0.805}	&	77.6\\
    \bottomrule
\end{tabular}
\end{center}
\end{table*}

%% file: files/conclusion.tex
In this paper, we have proposed a novel Bayesian deep learning based approach \emph{BayesBeat} for detection of Atrial Fibrillation from smartwatch Photoplethysmography (PPG) signals. \emph{BayesBeat} can effectively handle inherent input noise that makes it \textbf{robust} to motion artifacts. Our extensive experiments with the largest available dataset show that \emph{BayesBeat} outperforms the state-of-the-art PPG based AF detection approaches.
At the same time, \emph{BayesBeat} is significantly \textbf{lighter} since it requires much less number of parameters that makes it suitable for deployment in low-end mobile/wearable devices. Finally, \emph{BayesBeat} provides the uncertainty estimate of the prediction, which makes it \textbf{reliable} for use in real world setting and enables the \textbf{flexibility} to tune the model according to user needs. In future, we plan to deploy our model in commercially available smartwatches and test the real-time AF detection performance. Since the performance of Bayesian deep learning models also depends on the initially chosen prior, experimenting with different priors can be a promising future direction to achieve even better performance.

%% file: files/code_availability.tex
Trained weights and relevant source codes of \emph{BayesBeat} are publicly available at this github link:\\
\url{https://github.com/Sarathismg/BayesBeat}.

%% file: files/acknowledgement.tex
This research has been conducted at DataLab (datalab.buet.io), Department of CSE, Bangladesh University of Engineering and Technology (BUET), and supported by ICT Division, Government of the People's Republic of Bangladesh. We'd also like to thank
Dr. Md. Shamimur Rahman of National Heart Foundation of Bangladesh for his valuable suggestions and opinions about this work.